\pdfoutput=1

\documentclass[11pt]{article}
\usepackage[dvipsnames,svgnames,table]{xcolor}
\usepackage[preprint]{acl}

\usepackage{times}
\usepackage{latexsym}
\usepackage{CJKutf8}
\usepackage{tabularx}
\usepackage{booktabs}
\usepackage{babel}
\usepackage{listings}
\usepackage[size=tiny]{todonotes}

\usepackage[T1]{fontenc}

\usepackage[utf8]{inputenc}

\usepackage{microtype}

\usepackage{inconsolata}

\usepackage{graphicx}
\usepackage{amsmath}
\usepackage{amsfonts}
\usepackage{amssymb}

\newcommand{\se}[1]{\textcolor{black}{#1}}

\newcommand{\dl}[1]{\textcolor{black}{#1}}

\title{BatchGEMBA: Token-Efficient Machine Translation Evaluation with \\
Batched Prompting and Prompt Compression}

\author{
    Daniil Larionov\textsuperscript{1,2} \quad
    Steffen Eger\textsuperscript{1,3}\\
    \textsuperscript{1} NLLG, \textsuperscript{2} University of Mannheim, \textsuperscript{3} University of Technology Nuremberg\\
\href{mailto:daniil.larionov@uni-mannheim.de}{daniil.larionov@uni-mannheim.de}
}

\begin{document}
\maketitle
\setlength{\belowdisplayskip}{2pt} \setlength{\belowdisplayshortskip}{2pt}
\setlength{\abovedisplayskip}{2pt} \setlength{\abovedisplayshortskip}{2pt}

\begin{abstract}
Recent advancements in Large Language Model (LLM)-based Natural Language Generation evaluation have largely focused on single-example prompting, resulting in significant token overhead and computational inefficiencies. In this work, we introduce \textbf{BatchGEMBA-MQM}, a framework that integrates batched prompting with the GEMBA-MQM metric for machine translation evaluation. Our approach aggregates multiple translation examples into a single prompt, reducing token usage by 2--4 times (depending on the batch size) relative to single-example prompting. Furthermore, we propose a batching-aware prompt compression model that achieves an additional token reduction of 13--15\% on average while also showing ability to help mitigate batching-induced quality degradation. Evaluations across several LLMs (GPT-4o, GPT-4o-mini, Mistral Small, Phi4, and CommandR7B) and varying batch sizes reveal that while batching generally negatively affects quality (but sometimes not substantially), prompt compression does not degrade further, and in some cases, recovers quality loss. For instance, GPT-4o retains over 90\% of its baseline performance at a batch size of 4 when compression is applied, compared to a 44.6\% drop without compression. We plan to release our code and trained models at \url{https://github.com/NL2G/batchgemba} to support future research in this domain.
\end{abstract}

\section{Introduction}
\label{sec:introduction}

The rapid evolution of Natural Language Generation (NLG) evaluation metrics has led to significant improvements in aligning automated assessments with human judgments~\citep{papineni-etal-2002-bleu, popovic-2015-chrf, sun-etal-2022-bertscore, rei-etal-2020-comet}. 

\begin{figure}[h]
    \centering
    \includegraphics[width=\columnwidth]{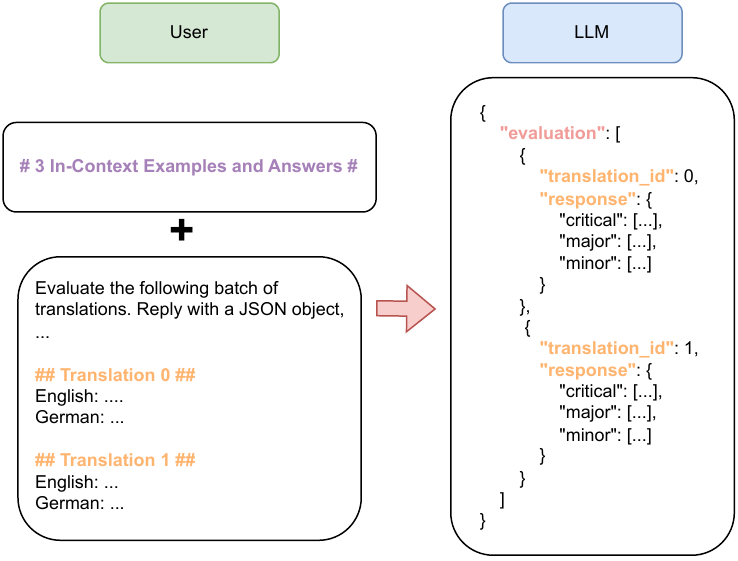}
    \caption{Overall Structure of BatchGEMBA-MQM prompt.}
    \label{fig:plot}
\end{figure}

Despite these advances, state-of-the-art methods such as GEMBA-MQM~\citep{kocmi-federmann-2023-gemba} have predominantly been applied in single-example settings, where each evaluation is conducted with a separate prompt. This approach, while effective, can be inefficient in terms of token usage and computational resources, especially when scaling to large datasets of multiple language pairs, web-scale data processing or online evaluation.

Batched prompting, which aggregates multiple examples into a single prompt, has recently emerged as a promising technique to substantially enhance efficiency by reducing the number of API calls and overall token consumption~\citep{cheng-etal-2023-batch}. However, to the best of our knowledge, no prompt-based metric for fine-grained evaluation of NLG has ever been evaluated in a batched prompting framework. In fact, GEMBA-MQM is posed to benefit the most from the token reduction in batching, since the majority of its tokens come from few-shot examples and detailed instructions, which are included into the prompts for each example. With batching, such components need only be included once per batch of \se{(say)} 2, 4, 8 or 16 examples, substantially improving token efficiency. Moreover, while prompt compression methods~\citep{pan-etal-2024-llmlingua, larionov2024promptoptme} have demonstrated their ability to reduce token count without sacrificing performance, their impact in a batched setting remains under-explored.

In this work, we explore the integration of batched prompting with \se{the} GEMBA-MQM prompt-based metric for machine translation evaluation and further investigate the effects of incorporating a batching-aware prompt compression strategy. Our research aims to address two key questions: (1) How does batched prompting affect the evaluation quality of prompt-based translation evaluation metrics, such as GEMBA-MQM, across various Large Language Models (LLMs)? (2) Can prompt compression enhance the efficiency of batched prompting without degrading the correlation with human quality judgments?

To answer these questions, we extend the conventional GEMBA-MQM prompt into a batched format, hereafter referred to as \emph{BatchGEMBA-MQM}, and systematically evaluate its performance both with and without the application of prompt compression. Our experiments span multiple batch sizes and assess a range of LLMs—including proprietary systems like GPT-4o and GPT-4o-mini as well as open-source models such as Mistral Small, Phi4, and CommandR7B, using Pearson-$r$ correlation with human judgement as a measure of metric quality and average token usage to assess efficiency.

The contributions of this paper are as follows:
\begin{itemize}
    \item We introduce the first application of batched prompting to machine translation evaluation, enabling simultaneous evaluation of multiple translation examples. The overall structure of our prompt is presented in Figure~\ref{fig:plot}.
    \item We propose a batching-aware prompt compression model that significantly reduces token usage while retaining essential evaluation cues.
    \item We conduct a comprehensive meta-evaluation across various LLMs and batch sizes, providing insights into the trade-offs between evaluation accuracy, token efficiency, and batch size.
    \item We offer an analysis of how prompt compression interacts with batched prompting, highlighting its potential to mitigate the token inefficiency inherent in existing prompting-based evaluation methods.
\end{itemize}

\section{Related Work}
\label{sec:related_work}

In this section we give an overview of existing approaches related to existing NLG evaluation metrics and approaches, prompt compression and batch prompting.

\subsection{NLG Evaluation Metrics}
Over the years, evaluation metrics have evolved from relying solely on surface-level comparisons to incorporating deeper semantic understanding. Traditional metrics such as BLEU~\citep{papineni-etal-2002-bleu} and chrF~\citep{popovic-2015-chrf} laid the foundation for automatic translation evaluation, but they are inadequate for nuanced semantic  comparisons.
More recently, embedding-based metrics like BERTScore~\citep{sun-etal-2022-bertscore} and learned metrics such as COMET~\citep{rei-etal-2020-comet} have shown a substantially higher alignment with human judgments by employing the capabilities of pretrained language models.

The prompting-based methods are of particular interest due to their few-shot capabilites. For instance, GEMBA-MQM \citep{kocmi-federmann-2023-gemba} employs a fixed three-shot prompting strategy with GPT-4 to detect error spans according to the Multidimensional Quality Metrics (MQM) framework, leading to state-of-the-art system ranking. Building on that, MQM-APE~\citep{lu-etal-2025-mqm} refines these error annotations by automatically post-editing translations, effectively filtering out non-impactful errors. Similarly, MQM-Chat~\citep{li-etal-2025-mqm} adapts the traditional MQM framework to the unique challenges of chat translation by introducing error categories tailored to capture ambiguity, buzzword issues, and dialogue inconsistency.

\dl{On the efficiency side, while model-based metrics deliver excellent quality, their high computational cost can be a barrier. EffEval~\citep{larionov-etal-2023-effeval} aimed to address the issue of computational cost in BERTScore and similar metrics by replacing large pretrained models with small ones. Their findings reveal that distilled models, such as TinyBERT, retains largest proportion of reference model quality while offering 27$\times$ speedup, compared to baseline.} COMETinho~\citep{rei-etal-2022-searching} and xCOMET-lite~\citep{larionov-etal-2024-xcomet} address this by distilling the original models into a much smaller, faster alternative that still retains most of their evaluative power. In connection to prompt-based metrics, \citet{larionov2024promptoptme} estimate that the original GEMBA-MQM is highly expensive to run --- it would take almost 1000 USD in API costs to fully evaluate 60k examples from WMT22 Metrics Challenge Dataset. In our work, we address the issue of inefficiency of prompt-based metrics through batched prompting and prompt compression.

PrExMe~\citep{leiter-eger-2024-prexme} is a large‐scale, systematic study that explores how different prompt templates affect the performance of open‐source LLM-based metrics for machine translation and summarization evaluation. In this work, over 720 prompt templates, spanning variations in base prompts, task descriptions, and output formats 
were evaluated across more than 6.6 million individual prompts. One key finding is that even seemingly minor changes (for instance, switching the output range from “0 to 100” to “-1 to +1”) can significantly alter metric rankings. 

While these metrics have considerably improved the alignment with human judgments, all prior studies evaluate them on single-instance prompts. In contrast, our work specifically examines how evaluation metrics perform when used with batched prompting, evaluating multiple examples at once.

\subsection{Prompt Compression}

The LLMLingua framework~\citep{jiang-etal-2023-llmlingua} revolutionized prompt compression through a three-stage coarse-to-fine methodology: (1) a dynamic budget controller allocating compression ratios across prompt components (instructions vs.\ demonstrations), (2) iterative token-level compression preserving conditional dependencies between lexical units, and (3) instruction-tuned distribution alignment between compressor and target LLMs.

Two more variants were built upon it. The first, LongLLMLingua~\citep{jiang-etal-2024-longllmlingua} introduces a prompt compression framework aimed at enhancing the efficiency and performance of large language models in long-context scenarios. Their approach leverages a question-aware coarse-to-fine compression strategy, combined with document reordering and adaptive compression ratios, to increase the density of key information while simultaneously reducing computational cost and latency. Experiments on benchmarks such as NaturalQuestions and LongBench demonstrate that this approach can boost performance by up to 21.4\% with substantially fewer tokens used.

LLMLingua-2~\citep{pan-etal-2024-llmlingua} is a task-agnostic prompt compression framework that leverages data distillation to enhance both efficiency and faithfulness. Their approach uses GPT-4 to generate an extractive text compression dataset and formulates compression as a token classification problem, ensuring that compressed prompts remain faithful to the original content. Experiments demonstrate that this approach reduces end-to-end latency by up to 2.9$\times$ at compression ratios of  2$\times$ to 5$\times$.

PromptOptMe~\citep{larionov2024promptoptme} introduces a novel two-stage approach combining supervised fine-tuning with preference optimization, achieving total token count reduction without quality degradation in machine translation evaluation tasks. This methodology \se{is} specifically trained to predict and preserve error spans as identified in MQM annotations through conditioned compression training, addressing an \se{important disadvantage} 
of prior task-agnostic compression methods like LLMLingua-2~\citep{pan-etal-2024-llmlingua}.

500xCompressor~\citep{li2024500xcompressor} introduces a substantially different approach to prompt compression by encoding extensive natural language prompts, up to 500 tokens, into as few as one token, achieving compression ratios from 6$\times$ to an impressive 480$\times$, leveraging the key--value cache mechanisms to capture fine-grained semantic details, mitigating both information loss and potential data leakage during evaluation. The experiments demonstrated that LLM retains up to 72\% of the original quality compared to using original uncompressed prompt. This approach demonstrates substantial performance boost, but it does require having access to the KV-cache for the respective models, which is not feasible for proprietary models, such as GPT-4o. In our work we aim to develop a general approach that would work with any LLM in a black-box fashion, without accessing its internals.

A key limitation of these methods is that they assume a single example per prompt. In our study, we investigate how prompt compression behaves when applied to batched inputs, revealing how token reduction and information preservation are affected.

\subsection{Batch Prompting}
Recent work by \citet{cheng-etal-2023-batch} introduces batch prompting as an efficient alternative to standard prompting. In contrast to the conventional approach, where each input sample requires a separate API call, batch prompting aggregates multiple samples into a single prompt. This method significantly reduces both token and time costs, as the authors theoretically show that token consumption decreases nearly inversely with the number of samples in each batch. Their extensive experiments on diverse benchmarks, including commonsense question answering, arithmetic reasoning, and natural language inference, demonstrate up to a 5$\times$ improvement in efficiency while maintaining or even improving downstream performance.

Although batch prompting has been shown to reduce both token and time costs, its interaction with prompt compression remains unexplored. On top of that, existing work only explores trivial output formatting for QA tasks, where each example prediction is only one sentence, while \se{MQM-based MT evaluation} strongly depend\se{s} on fine-grained error identification and categorization, which requires more complex structured output format. Such complex output formatting appears to be especially challenging for LLMs to comply with, given the need to produce responses for all examples in a batch at-once. Our work explores the performance of several different LLMs in this batched evaluation setting, for multiple batch sizes and with prompt compression.

\section{Method}

This section describes our approach to prompt compression for batched prompting in the evaluation of machine translation quality. Our method adapts the GEMBA-MQM prompt—originally designed for single-example few-shot evaluation—into a batched format that enables the evaluation of multiple translation examples within a single prompt. We call this adapted prompt ``BatchGEMBA-MQM''. The original GEMBA-MQM prompt is token-intensive, which makes it less efficient for use with target LLMs. To address this, we extend the ideas introduced by PromptOptMe, by creating a batching-aware prompt compression model  which is able to compress batches of examples so that only the essential error-relevant information is retained, thereby creating more compact input prompts for target LLMs. 

\subsection{Prompt}
First, in our adaptation we transform the original GEMBA-MQM prompt into a token-efficient format that can handle a batch of examples simultaneously, while preserving few-shot learning capability. Our approach for prompt adaptation extends on~\citet{cheng-etal-2023-batch}. In the first part of the prompt, we insert a first mini-batch of examples which consists of 3 in-context examples from the original GEMBA-MQM prompt, followed by JSON-formatted evaluations of those examples. This demonstration, along with few-shot learning guidance, steers the model to follow pre-defined output format. It is especially important for batched evaluation, as it requires clearly distinct responses for each individual example from the batch. We choose JSON-format over the plain-text one to enable the use of guided generation methods~\citep{willard2023efficient}, which further helps minimize inconsistencies in model outputs.

\subsection{Training}
The training process for our batching-aware compression model follows a two-stage approach inspired by PromptOptMe, with specific modifications to support batched prompting. In the first stage, we generate extensive training data through a combination of data collection and controlled random token removal. We begin by gathering batches of examples of varying sizes (from 2 to 8 examples per batch) from the WMT Metrics Challenge dataset~\citep{freitag-etal-2023-results}. Each example within a batch comprises a source segment and a translation. To simulate compression, we apply random token removal to each batch, eliminating a random proportion (from 1\% to 70\% of segment length) of tokens while preserving error-relevant spans. For this, we extract human-created annotations of error spans within each translation from the dataset and use them to condition the token removal algorithm. This strategy encourages the compression model to focus on retaining the most crucial information even when significant portions of text are removed.

In the second stage, we utilize the pretrained compression model from Stage 1 to generate compressed batches examples from batches derived from the WMT MQM dataset. For each unique batch, multiple compressed versions are produced using sampling decoding with random seeds. A filtering is then applied to discard any compressions that fail to meet the defined batched format, which would prevent us from processing them further. Each remaining compression is evaluated by submitting it as a prompt to a target LLM, whose evaluation of translation quality is compared against human-produced assessments. For each batch, we determine the best and worst compressions based on this evaluation. In instances where multiple compressions achieve similar evaluation quality, the one with the fewest tokens is selected to ensure optimal compression efficiency.

\section{Experiments}
\label{sec:experiments}
In this section, we describe the experimental setup used for training the compressor LLM as well as for the meta-evaluation of batched BatchGEMBA-MQM at various batch sizes, both with and without prompt compression.

\subsection{Model Training}
\label{sec:experiments/model-training}
We follow the methodology of PromptOptMe in selecting the dataset for training. Specifically, we used a portion of the WMT Metrics Challenge dataset with MQM annotations spanning the years 2020–2022. No specific subset or domain was excluded for validation; instead, we employ a randomly selected held-out set comprising approximately 10\% of the dataset at each stage for validation and early stopping.

In the first stage of training, we generate random compressions with varying compression rates and batch sizes. Generation is limited to 50k examples, with batches created using examples from a single language pair only. Batch lengths and compression rates are uniformly sampled. We use Llama 3.2 3B ~\citep{dubey2024llama} 
as base model for our fine-tuning. We train the model for one epoch using LoRA~\citep{hulora} adapters with $r = 64$ and $\alpha = 16$, resulting in 97M trainable parameters out of a total of 3.3B. The training is performed with the AdamW~\citep{loshchilov2019decoupledweightdecayregularization} optimizer, using a learning rate of 5e-5, a warmup phase covering 0.06\% of the training steps, and linear decay thereafter.

In the second stage, we generate a dataset of compression preferences for approximately 10k batches. We employ the same model as PromptOptMe---LLaMa 3.2 3B. 
For preference optimization, we apply a pipeline similar to that of PromptOptMe. We retain the LoRA adapter from stage 1 and fine-tune it further using the ORPO~\citep{hong2024reference} preference optimization algorithm, while keeping the hyperparameters identical to those used in stage 1, as they yielded the best results during development runs.

\subsection{Meta-Evaluation}
\label{sec:experiments/meta-eval}
In the meta-evaluation, we assess the performance of each LLM under batched prompting at various batch sizes and investigate the effect of prompt compression produced by the model trained in Section~\ref{sec:experiments/model-training}. For this purpose, we evaluate  several LLMs—both proprietary and open-source:
\begin{itemize}
    \item \textbf{GPT-4o} and \textbf{GPT-4o-mini}~\citep{openai2024gpt4ocard}: proprietary models from OpenAI.
    \item \textbf{Phi4}~\citep{abdin2024phi}: a small open LLM trained on a mix of synthetic and non-synthetic data.
    \item \textbf{Mistral Small}\footnote{\url{https://huggingface.co/mistralai/Mistral-Small-24B-Instruct-2501}}: an open LLM with 24B parameters.
    \item \textbf{CommandR7B}\footnote{\url{https://huggingface.co/CohereForAI/c4ai-command-r7b-12-2024}}: an open LLM trained on a mixture of 23 languages.
\end{itemize}

Each of these models was tested with batch sizes of 1, 2, 4, 8, and 16. A batch size of 16, not present during training, was introduced to assess generalization to higher batch sizes. We use the WMT23 Metrics Challenge Dataset~\citep{freitag-etal-2023-results}, which comprises data points for three language pairs: English–German, Hebrew–English, and Chinese–English. Notably, the Hebrew–English language pair was absent from the training dataset. Furthermore, while the English–German subset of the WMT23 Metrics dataset consists of paragraph-level examples, our compression model was primarily exposed to sentence-level translations during training.


\subsection{Experimental Details}
We assess the performance of batched prompting using the following metrics:
\begin{itemize}
    \item \textbf{Pearson Correlation:} between BatchGEMBA-MQM evaluations of 
    examples and respective human judgments.
    \item \textbf{Token Usage:} total tokens consumed, averaged across language pairs.
    \item \textbf{Error Rate:} percentage of examples with misformatted LLM outputs.
\end{itemize}

Model training experiments were conducted on a node of university cluster, which had 4$\times$RTX A5000 GPU. In total, the model training runs consumed 128 GPU-hours.

\section{Results}
\label{sec:results}

\begin{table*}[h!]
    \centering
    \begin{tabular}{llrrrrr}
    \hline
     Model         & Comp.        &     1 &     2 &     4 &     8 &    16 \\
    \hline
     GPT-4o (only En-De)       & - & 0.613 & 0.340 & 0.384 & 0.269 & 0.015 \\
                   & + & 0.603 & 0.498 & 0.548 & 0.189 & 0.028 \\
     GPT-4o-mini   & - & 0.455 & 0.342 & 0.322 & 0.342 & 0.304 \\
                   & + & 0.494 & 0.326 & 0.361 & 0.320 & 0.303 \\
     Mistral Small & - & 0.271 & 0.075 & 0.078 & 0.039 & 0.014 \\
                   & + & 0.328 & 0.087 & 0.058 & 0.040 & 0.023 \\
     Phi4          & - & 0.349 & 0.150 & 0.165 & 0.179 & 0.148 \\
                   & + & 0.382 & 0.137 & 0.135 & 0.183 & 0.163 \\
     CommandR7B    & - & 0.233 & 0.105 & 0.107 & 0.084 & 0.055 \\
                   & + & 0.154 & 0.107 & 0.076 & 0.084 & 0.043 \\
    \hline
    \end{tabular}
    \caption{Meta-evaluation of BatchGEMBA-MQM with and without prompt compression on the WMT23 Metrics Challenge dataset. Batch sizes of 1, 2, 4, 8, and 16 indicate the number of examples included in each prompt.}
    \label{tab:correlations}
\end{table*}

Table~\ref{tab:correlations} reports the Pearson $r$ correlations between BatchGEMBA-MQM evaluations and human judgments for several backbone LLMs. When increasing the batch size from 1 to 2 and 4 examples, all models experience a decline in correlation values, though the magnitude of the reduction varies. For instance, in the non-compressed condition, GPT-4o correlation decreases from 0.613 at batch size 1 to 0.340 at batch size 2 (a drop of approximately 44.6\%), while GPT-4o-mini falls from 0.455 to 0.342 (a 24.6\% reduction). Surprisingly, under prompt compression, GPT-4o retains a higher fraction of its initial performance, dropping from 0.603 at batch size 1 to 0.498 at batch size 2 (17.4\% reduction), with GPT-4o-mini showing a similar trend (from 0.494 to 0.326). Mistral Small and Phi4 exhibit lower initial correlations that decrease further with increasing batch size, and CommandR7B, shows a reduction from 0.233 to 0.105 in the non-compressed mode and from 0.154 to 0.107 under compression. These numbers demonstrate that while all models are adversely affected by increasing batch size, the extent of the decline, and the surprising benefit of compression, differs across systems. Due to high experiment cost, here and afterwards we only report scores for En-De for GPT-4o model.

\begin{figure}
    \centering
    \includegraphics[width=\columnwidth]{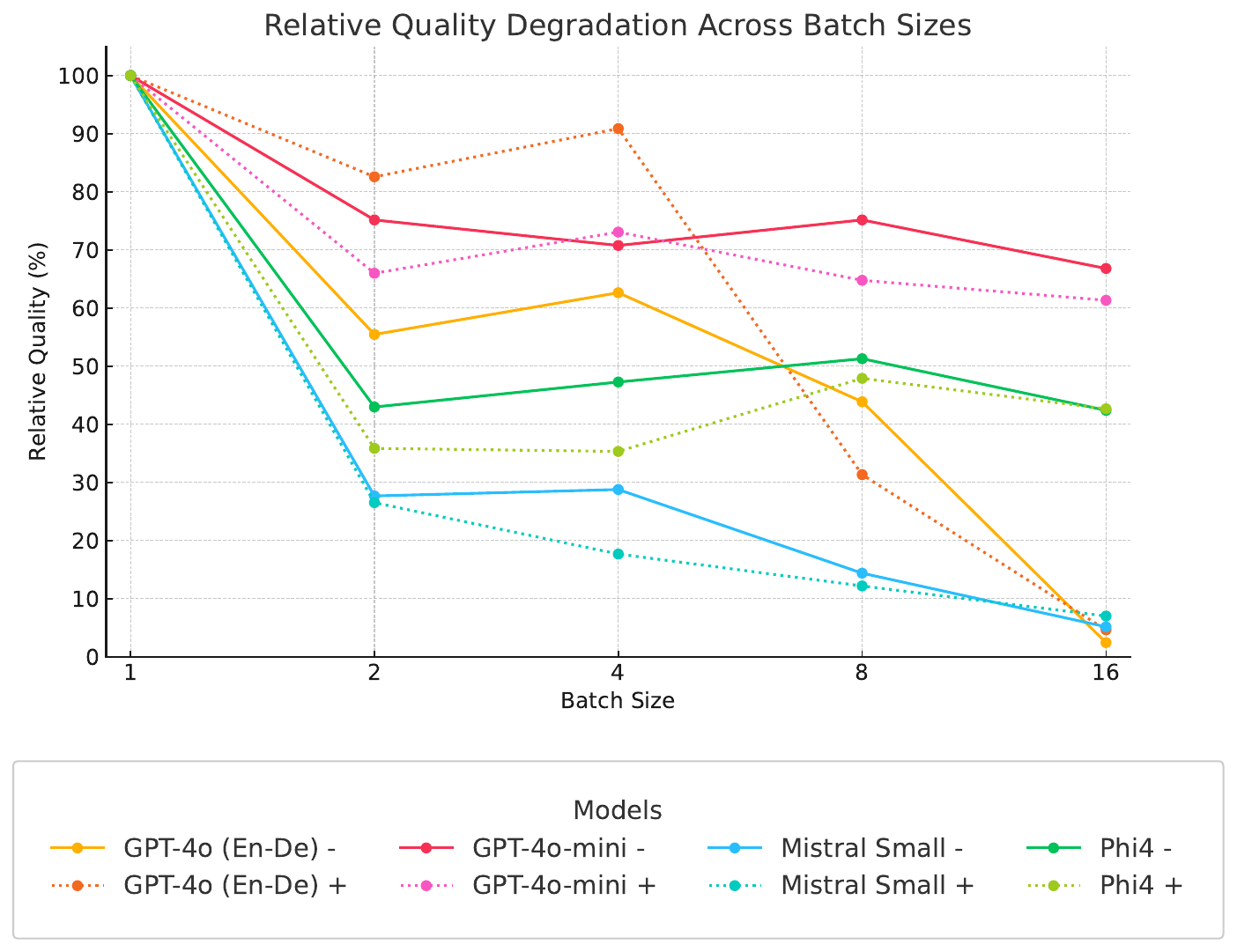}
    \caption{Relative quality degradation, normalized to the each models performance at batch size $1$. `-' in model name indicate evaluation without compressed prompts, `+' indicates compressed prompts.}
    \label{fig:illustration}
\end{figure}

\se{Figure \ref{fig:illustration} shows similar information as Table \ref{tab:correlations} but reports metric quality as a function of batch size, normalized by the quality at batch size $1$. We observe the following: (i) All models experience a quality drop at batch sizes larger than $1$. (ii) At batch size $2$, only the GPT-4o family retains more than 50\% of its quality attained at batch size $1$. (ii) The same is true for batch size $4$, where surprisingly 3 out of 4 GPT-4o model variants perform better than at batch size $2$, retaining more than 60\% of their batch size $1$ quality. The compressed version of GPT-4o even retains more than 90\%. (iii) Somewhat surprisingly, GPT-4o-mini remains stable for larger batch sizes, while GPT-4o drops dramatically, achieving only 2\% of its batch size 1 quality at batch size 16. (iv) Both Mistral and Phi4 do generally not gracefully degrade, losing more than 50\% of their batch size $1$ quality throughout. (v) Batching and compression do not systematically result in lower quality compared to batching alone. Instead, sometimes the batched and compressed model variant performs better than the one which only undergoes batching.}

\begin{table*}[h!]
    \centering
    \begin{tabular}{llrrrrr}
    \hline
     Model         & Comp.        &     1   &     2   &     4   &     8   &    16 \\
    \hline
     GPT-4o (only En-De)       & - & 5.4M    & 4.1M    & 2.8M    & 2.1M    & 1.7M \\
                   & + & 4.7M    & 3.8M    & 2.6M    & 1.9M    & 1.7M \\
     GPT-4o-mini   & - & 13.6M   & 9.9M    & 5.9M    & 3.9M    & 2.9M \\
                   & + & 11.6M   & 8.9M    & 5.4M    & 3.7M    & 2.8M \\
     Mistral Small & - & 14.0M   & 10.1M   & 6.0M    & 3.9M    & 2.9M \\
                   & + & 11.9M   & 9.1M    & 5.5M    & 3.7M    & 2.8M \\
     Phi4          & - & 13.7M   & 10.2M   & 6.5M    & 4.4M    & 3.3M \\
                   & + & 12.8M   & 9.6M    & 5.9M    & 4.2M    & 3.2M \\
     CommandR7B    & - & 12.5M   & 9.9M    & 5.9M    & 3.8M    & 2.8M \\
                   & + & 10.4M    & 8.9M    & 5.4M    & 3.5M    & 2.7M \\
    \hline
    \end{tabular}
    \caption{Total token usage for BatchGEMBA-MQM evaluations with and without prompt compression. Values are averaged across language pairs for batch sizes of 1, 2, 4, 8, and 16 examples.}
    \label{tab:token_usage}
\end{table*}

Table~\ref{tab:token_usage} details the total token usage, averaged across language pairs. In general, token usage decreases as the batch size increases. For instance, GPT-4o in the non-compressed condition uses 5.4M tokens at batch size 1, which reduces to 4.1M at batch size 2 and further to 2.8M at batch size 4. GPT-4o-mini similarly shows a decline from 13.6M to 9.9M and then to 5.9M tokens over the same transition. Mistral Small and Phi4 follow similar trends, while CommandR7B, reduces from 10.1M tokens at batch size 1 to 9.9M at batch size 2 and 5.9M at batch size 4. \se{Compression on top of batching yields a small additional efficiency boost. For instance, while GPT-4o-mini has a total token usage of 13.6M at batch size $1$, it has 9.9M at batch size $2$ and the compressed version has 8.9M.} 

\begin{table*}[ht]
    \centering
    \begin{tabular}{llrrrrr}
    \hline
     Model         & Comp.        &     1    &     2    &     4    &     8    &    16 \\
    \hline
     GPT-4o (only En-De)        & - & 0.0\%   & 0.2\%  & 0.3\%  & 0.0\%  & 0.2\% \\
                   & + & 0.0\%   & 0.1\%  & 0.1\%  & 4.1\%  & 0.0\% \\
     GPT-4o-mini   & - & 0.0\%   & 0.0\%  & 0.0\%  & 0.0\%  & 0.0\% \\
                   & + & 0.0\%   & 0.0\%  & 0.0\%  & 0.0\%  & 0.2\% \\
     Mistral Small & - & 0.0\%   & 0.2\%  & 0.4\%  & 0.4\%  & 0.5\% \\
                   & + & 0.0\%   & 0.2\%  & 0.4\%  & 0.5\%  & 0.8\% \\
     Phi4          & - & 13.3\%  & 6.3\%  & 4.9\%  & 5.5\%  & 7.1\% \\
                   & + & 1.6\%   & 3.8\%  & 4.6\%  & 5.2\%  & 6.6\% \\
     CommandR7B    & - & 16.7\%  & 0.1\%  & 0.1\%  & 0.1\%  & 0.1\% \\
                   & + & 30.5\%  & 0.1\%  & 0.2\%  & 0.2\%  & 0.1\% \\
    \hline
    \end{tabular}
    \caption{Error rates for BatchGEMBA-MQM evaluations with and without prompt compression. Error rates are averaged across language pairs for batch sizes of 1, 2, 4, 8, and 16 examples.}
    \label{tab:error_rate}
\end{table*}

Table~\ref{tab:error_rate} presents the error rates, also averaged across language pairs. Error rates are calculated as a percentage of examples for which the evaluation was not obtained due to misformatted output of the LLM. For instance, GPT-4o in the non-compressed setting has an error rate of 0.0\% at batch size 1, which increases slightly to 0.2--0.3\% at batch sizes 2 and 4; its compressed variant maintains 0.0\% error for smaller batches before rising to 4.1\% at batch size 8. GPT-4o-mini consistently reports near-zero error rates across configurations. Mistral Small and Phi4 exhibit modest error rates that vary with batch size, whereas CommandR7B, shows an unusually high error rate of 16.7\% at batch size 1 in the non-compressed mode. Surprisingly, that drops substantially to 0.1\% for larger batches, with its compressed version starting at 30.5\% and converging to 0.1--0.2\% as the batch size increases. 

\section{Discussion}
\label{sec:discussion}

Our results provide several insights into the use of batched prompts in machine translation evaluation, the impact of prompt compression, and the inherent challenges associated with batching.

Regarding the ability of LLMs to handle batched prompts, the experiments reveal a mixed picture. Several models exhibit a notable decline in correlation values when moving from single-example prompts to small batches, indicating that batching can stress the evaluation capabilities of certain systems. For instance, \se{as discussed}, GPT-4o and GPT-4o-mini show moderate reductions in performance. GPT-4o at batch-size 2 with compression applied retains 81.3\% of the baseline (single-example, uncompressed) metric quality, \se{at batch size 4 it even retains more than 90\%}. Other models, such as Mistral Small, suffer steep drops. This suggests that the architecture and training of each LLM play a significant role in determining its robustness to increased context size and inter-example interference. In effect, some LLMs can integrate multiple examples without severe degradation in quality, whereas others are not suited to batch processing.

The effect of the prompt compression method is also of particular interest. Our findings indicate that, in most cases, the evaluation performance with compressed prompts remains very close to that of the original, non-compressed prompts. This suggests that our compression method successfully retains the critical information necessary for quality evaluation while reducing token usage. Notably, in several instances, the compressed prompts yield slightly better correlation scores than their non-compressed counterparts. These improvements, though unexpected, can be interpreted as a beneficial side effect of reducing redundant information that might otherwise adversely affect the model. Overall, the compression method appears to offer a valuable trade-off, maintaining evaluation accuracy while improving token efficiency.

Finally, our study highlights several challenges inherent to batching. As batch sizes increase, the interference among multiple examples in a single prompt may lead to decreased performance, as seen in the significant correlation drops for some models. Additionally, the balance between token reduction and evaluation quality becomes more critical with larger batches, particularly when the model's context window is limited. A careful calibration of batch size for each LLM is necessary, as overly large batches may overwhelm the model and compromise the quality of the evaluation. Therefore, while batching offers clear advantages in terms of token efficiency, it also introduces challenges that must be addressed through model-specific tuning and further investigation.

\section{Conclusion}
\label{sec:conclusion}

In this work, we extended the GEMBA-MQM prompt into a batched prompting approach, BatchGEMBA-MQM, to tackle the inefficiencies inherent in single-example LLM-based evaluations. Our goal was to reduce token usage and computational overhead while maintaining a close alignment with human judgments. Our results show that the impact of batching is highly dependent on the underlying LLM: while models such as GPT-4o and GPT-4o-mini experience only moderate performance declines with increased batch sizes, others like Mistral Small are substantially more affected. \se{This is an intriguing unexpected finding.}

Notably, the prompt compression has been demonstrated to preserve essential evaluation signals. In several configurations, the compressed prompts not only matched the performance of their uncompressed counterparts but even yielded a modest improvement. This indicates that a well-designed compression strategy can help alleviate inter-example interference by filtering out redundant information.

However, the challenges associated with batch scaling remain significant. As batch sizes increase, a balance must be maintained between token efficiency and evaluation accuracy. In future work, we plan to expand training to include more languages, different evaluation tasks, as well as to incorporate novel reinforcement learning techniques into the second stage of our training pipeline. For instance, we plan to adapt Group-Relative Policy Optimization (GRPO)~\citep{shao2024deepseekmath} strategy, with carefully designed reward function combination, which would allow us simultaneously incorporate efficiency component, format guidance and the preservation of evaluation quality.

Overall, our study provides a pathway towards more resource-efficient and scalable LLM-based evaluation for machine translation and potentially other NLG tasks.


\section*{Limitations}
\label{sec:limitations}

Several limitations of our work warrant discussion. First, our evaluation framework has been focused only on machine translation, leaving its applicability to other tasks, such as summarization evaluation or dialogue evaluation, unexplored. Further work could include more evaluation tasks. Second, due to cost constraints, our testing on GPT-4o, which appears to be one of the most batching-capable among the evaluated models, was limited. Further work should explore the performance of GPT-4o and similar models more thoroughly.

Additionally, our training did not incorporate more variability in batch sizes, which could have further boosted evaluation quality. We also observed that the compression rate tends to decrease as batch size increases. This behavior likely stems from our preference tuning, which prioritized preserving evaluation quality over minimizing token count. Future work could address this by adapting a GRPO strategy with a mixture of reward functions that simultaneously targets quality preservation, output format compliance, and token reduction.

For the CommandR7B model, we note an unusually high error-rate. This might be the result of suboptimal prompt design in the case of this model. Further research could address this by evaluating multiple prompt designs for each model.

Finally, our investigation into low-resource language pairs is limited to Hebrew--English. Broadening the scope to include additional low-resource languages would help assess the robustness of the proposed approach.

\bibliography{anthology,custom}

\appendix

\section{Additional Details}

\subsection{Potential Risks}

The use of batched prompting may introduce errors in translation evaluation. Therefore, practitioners should be careful in applying this approach in high-risk domains, such as medical or legal texts.

\subsection{Artifact Use}
We use Llama 3.2 model, released by Meta, as a base model for training our batched prompt compression model. Our use of this model is in compliance with the Acceptable Use Policy and Llama 3.2 Community License.

\subsection{Software versions}
We use the following software packages.
\begin{itemize}
    \item Axolotl v0.6.0
    \item PyTorch v2.4
    \item TRL v0.12.2
\end{itemize}

\end{document}